\documentclass[runningheads]{llncs}
\usepackage[T1]{fontenc}
\usepackage{graphicx}
\usepackage{booktabs}  
\usepackage{xcolor}
\usepackage{cite}
\usepackage{array}     
\usepackage{marvosym}
\usepackage{adjustbox} 
\usepackage{multirow}  
\usepackage{float}     
\usepackage{tabularx}  
\usepackage{ragged2e}  
\usepackage{hyperref}
\usepackage{CJKutf8}
\begin{document}
\begin{CJK*}{UTF8}{gbsn}

\title{Identifying Speakers and Addressees of Quotations in Novels with Prompt Learning}
\author{Yuchen Yan \and Hanjie Zhao \and Senbin Zhu\and Hongde Liu\and Zhihong Zhang\and Yuxiang Jia\textsuperscript{(\Letter)}}
\authorrunning{Y. Yan et al.}
\institute{School of Computer and Artificial Intelligence, Zhengzhou University, Zhengzhou, P.R.China\\
\email{\{yanyuchen,hjzhao\_zzu,nlpbin,lhd\_1013\}@gs.zzu.edu.cn},\\
\email{\{iezhzhang,ieyxjia\}@zzu.edu.cn}}
%
\maketitle              
\begin{abstract}
Quotations in literary works, especially novels, are important to create characters, reflect character relationships, and drive plot development. Current research on quotation extraction in novels primarily focuses on quotation attribution, i.e., identifying the speaker of the quotation. However, the addressee of the quotation is also important to construct the relationship between the speaker and the addressee. To tackle the problem of dataset scarcity, we annotate the first Chinese quotation corpus with elements including speaker, addressee, speaking mode and linguistic cue. We propose prompt learning-based methods for speaker and addressee identification based on fine-tuned pre-trained models. Experiments on both Chinese and English datasets show the effectiveness of the proposed methods, which outperform methods based on zero-shot and few-shot large language models.

\end{abstract}

\section{Introduction}
Quotations and dialogues are crucial components of literary works, especially novels\cite{page1988speech}. Quotes advance the plot and provide information about personality, emotion and social relationship of characters. Therefore, identifying the speaker and addressee of each quotation is essential to uncover the latent character traits and relationships within dialogue. 

\begin{figure}
  \centering
  \includegraphics[width=0.9\linewidth]{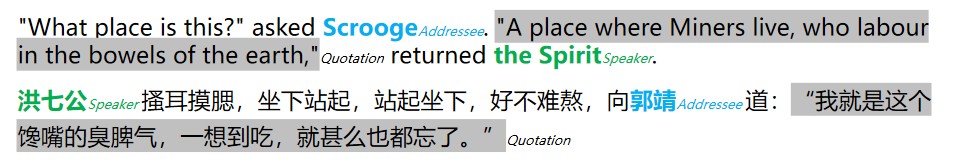}
  \caption{Examples of Quotation Speaker and Addressee}
  \label{fig1}
\end{figure}

We aim to identify the speaker and addressee of a given quotation within its context. As shown in Figure \hyperref[fig1]{1}, we obtain the \{speaker, quotation, addressee\} triplet from the text, which serves as the basis for further research on the implicit character relationships in the dialogue text. Current research on novel dialogues and quotations primarily focuses on quotation attribution, i.e., speaker identification, with insufficient attention given to addressee identification. Major novel quotation corpora, such as RiQua\cite{papay2020riqua} and PDNC\cite{vishnubhotla2022project}, include addressee elements but have not been extensively studied in this regard. Additionally, there is currently no Chinese corpus that includes addressee elements. To address this gap, based on JY-Quote\cite{xie2023corpus}, we annotate JY-QuotePlus, a corpus that contains four quotation elements, i.e., speaker, addressee, speaking mode and linguistic cue. To identify speaker and addressee of quotations, we employ a machine reading comprehension (MRC) approach, fine-tuning pre-trained models T5 and PromptCLUE\footnote{https://github.com/clue-ai/PromptCLUE} for English and Chinese novels respectively. We also assess the performance of general large language models on this task, and comparative results demonstrate the effectiveness of our approach.

The main contributions of this paper are summarized as follows:

\begin{enumerate}
\item[$\bullet$]We propose the first Chinese corpus containing both speakers and addressees to facilitate the research of speakers and addressees identification of quotations in Chinese novels, and perform a detailed analysis of the corpus. The corpus will be made publicly available at https://github.com/LimboChen/JY-QuotePlus.
\item[$\bullet$]We propose fine-tuned pre-trained models (PTMs) based on Transformer architecture, and design prompts to identify speakers and addressees of quotations. As a comparison, we select two popular large language models (LLMs) as baselines to evaluate the performance of LLMs on this task.
\item[$\bullet$]Experiments are conducted on both our Chinese corpus JY-QuotePlus and and English corpus RiQua. Experimental results demonstrate the effectiveness and superiority of the proposed fine-tuned PTMs over few-shot LLMs.
\end{enumerate}

\section{Related Work} 
The advancement of natural language processing (NLP) technologies has significantly accelerated the automatic analysis of literary works, covering various aspects of literary research\cite{JiaYuxiang2022novelcharacter}. Given the importance of quotations and dialogues in novels, research on quotations within novel texts is particularly significant.

There are several publicly available datasets with various types of quotation elements for quotation extraction tasks, primarily in rich-resource languages such as English and Chinese. Table \hyperref[tab1]{1} provides an overview of these corpora. However, there is currently no Chinese corpus that includes both speakers and addressees.

Speaker identification, also known as quotation attribution, is a classic task in the field of NLP, with significant applications in both literary  and news domains. Elson et al.\cite{mckeown2010automatic} utilized rule-based and statistical learning based methods. He et al.\cite{he2013identification} were the first to use  machine learning approach to attribute quotations to speaker entities. Cuesta-Lazaro et al.\cite{cuesta2022does} developed a pipeline for novel quotation attribution, which included quotation identification, character and alias recognition, and dialogue attribution, utilizing a deep learning model based on dialogue state tracking for speaker attribution. Vishnubhotla et al.\cite{vishnubhotla2023improving} also decomposed the quotation attribution task into similar subtasks, improving BookNLP\cite{sims2020measuring} to limit the candidate set and resolve mention spans in the coreference step, and directly linking quotations to entities. However, the pipeline approach faces challenges due to the lack of perfect solutions for each subtask, leading to error accumulation affecting the final results. In contrast, Yu et al.\cite{yu2022end} proposed the first end-to-end (E2E) speaker identification system, utilizing a pre-trained language model to construct an MRC model that attributed quotations to mention-level speakers.
\begin{table}
\centering
\caption{Quotation Annotation Corpora of Novels}
\label{tab1}
\begin{tabular}{>{\centering\arraybackslash}p{2cm} >{\centering\arraybackslash}p{1.5cm} >{\centering\arraybackslash}p{2cm} >{\centering\arraybackslash}p{3cm} >{\centering\arraybackslash}p{3cm}}
    \toprule
    Corpus & Language & \#Quotes & Quotation Elements & Character Type \\
    \toprule
    CQSAC\cite{mckeown2010automatic} & English & 3,176 & Speaker & mention \\
    QuoteLi3\cite{muzny2017two} & English & 3,106 & Speaker & entity,mention \\
    LitBank\cite{bamman2019annotated} & English & 1,765 & Speaker & entity \\
    RiQua\cite{papay2020riqua} & English & 5,963 & Speaker,Cue,Addressee & mention \\
    PDNC\cite{vishnubhotla2022project} & English & 35,978 & Speaker,Addressee & entity \\
    \toprule
    WP\cite{chen2019chinese} & Chinese & 2,548 & Speaker & entity \\
    JY\cite{jia2020speaker} & Chinese & 31,733 & Speaker & entity \\
    CSI\cite{yu2022end} & Chinese & 65,540 & Speaker & entity \\
    JY-Quote\cite{xie2023corpus} & Chinese & 31,992 & Speaker,Mode,Cue & mention \\
    \bottomrule
\end{tabular}
\end{table}

Addressees, as another crucial participant in dialogue, often appear implicitly in the dialogue context, which makes their identification more challenging than the speakers. Yeung et al.\cite{yeung2017identifying} pioneered the specific identification of addressees in dialogue, using a Conditional Random Field (CRF) sequence labeling algorithm to identify addressees within the context. Ek et al.\cite{ek2018identifying} employed a sequence labeling method applied to a given set of characters, selecting the most likely characters from the current chapter's character list based on features related to the current dialogue, previous narrative, and the full preceding context.

\section{Corpus Construction}
\subsection{Annotation Process}
To address the current lack of addressee element in Chinese corpora, we annotate the addressees of quotations based on the existing JY\cite{jia2020speaker} dataset and JY-Quote\cite{xie2023corpus} corpus. This results in JY-QuotePlus, the most comprehensive Chinese quotation corpus to our knowledge, containing a full set of quotation elements. The annotation of addressees is iteratively performed by four graduate students and involves the following steps.

We firstly formulate the initial annotation guidelines after studying the dataset. Each annotator then performs a small-scale pilot annotation. Based on the issues encountered, we discuss and refine the guidelines. In the formal annotation phase, the first annotator performs the initial annotation and records any issues. The second annotator reviews these annotations, and any discrepancies are resolved through discussion until consensus is reached. The annotation is carried out on a platform developed by our laboratory.

\subsection{Annotation Schema and Guidelines}

The addressee annotation is taken from a part of the JY corpus\cite{jia2020speaker}, namely \textit{The Legend of the Condor Heroes} (射雕英雄传), and is annotated based on the text segmentation of this corpus. Each annotated text segment included four parts: the quotation requiring addressee identification, the five preceding and five following sentences of the quotation, and the character entities mentioned in these contexts as candidates. Annotators are required to identify and annotate the addressee of the specified quotation, following guidelines below:
\begin{enumerate}
\item[$\bullet$]Annotate Only Candidates from the List: When a clear and complete addressee entity from the candidate list appears close to the quotation, it should be annotated as the addressee.
\item[$\bullet$]Priority Order: In general, prioritize context from the preceding text over the following text.
\item[$\bullet$]Immediate Addressee Identification: In cases where the addressee appears immediately after the quotation (often in two-person dialogues), directly annotate the subsequent addressee.
\item[$\bullet$]Avoid Quotation-Internal Addressees: Try not to annotate addressees mentioned within the quotation itself, unless no other addressees are identified in the context.
\item[$\bullet$]Single Annotation per Character: Each character should be annotated only once.
\item[$\bullet$]Multiple Addressees in Group Dialogues: If the quotation is directed at multiple people in a group setting without specifying one individual, annotate all characters present as addressees.
\item[$\bullet$]No Addressee Scenarios: Do not annotate addressee in cases of internal monologue, self-talk, or when there is no response.
\end{enumerate}	

\subsection{Corpus Analysis}
To validate the quality of annotations, we conduct an inter-annotator agreement (IAA) evaluation between annotators. The consistency of annotations in this study is assessed using the F1 score and Cohen's Kappa score. The F1 score for addressee annotation is 98.78\%, while the Kappa score is 0.9717, indicating a high reliability of the addressee information in this corpus.

\begin{table}
\centering
\caption{Corpus Statistics (8,144 quotations)}
\label{tab2}
\begin{tabular}{>{\centering\arraybackslash}p{3cm} >{\centering\arraybackslash}p{2.5cm} >{\centering\arraybackslash}p{2.5cm} >{\centering\arraybackslash}p{1.5cm} >{\centering\arraybackslash}p{1.5cm}}
    \toprule
    Metrics & Speaker & Addressee & Cue & Mode \\
    \toprule
    Occurrence Rates & 100\% & 100\% & 99.13\% & 48.01\% \\
    IAA(F1) & 99.21\% & 98.78\% & 99.07\% & 94.06\% \\

    \bottomrule
\end{tabular}
\end{table}

By combining the newly annotated addressee information with existing quotation elements in JY-Quote\cite{xie2023corpus}, we construct JY-QuotePlus. To facilitate effective use of this corpus in subsequent tasks, we ensure that each quotation includes both speaker and addressee entities. The overall statistics for the corpus elements are presented in Table \hyperref[tab2]{2}. Out of the 8,144 annotated quotations, 99.13\% include cues and 48.01\% include the mode of speech. The Cue element primarily expresses or implies the occurrence of a verbal event, while the Mode element primarily indicates how the speaker expresses the quotation in terms of emotions, tone, or intonation. Additionally, we also refer to the annotation consistency of other quotation elements\cite{xie2023corpus}. The consistency values for all these elements exceed 90\%, indicating the reliability of the dataset.

In the field of automated literary analysis, constructing virtual social relationship networks for characters is a significant research direction. Dialogue relationships represent one of the most straightforward and revealing forms of social connections between characters.
\begin{figure}
  \centering
  \includegraphics[width=\linewidth]{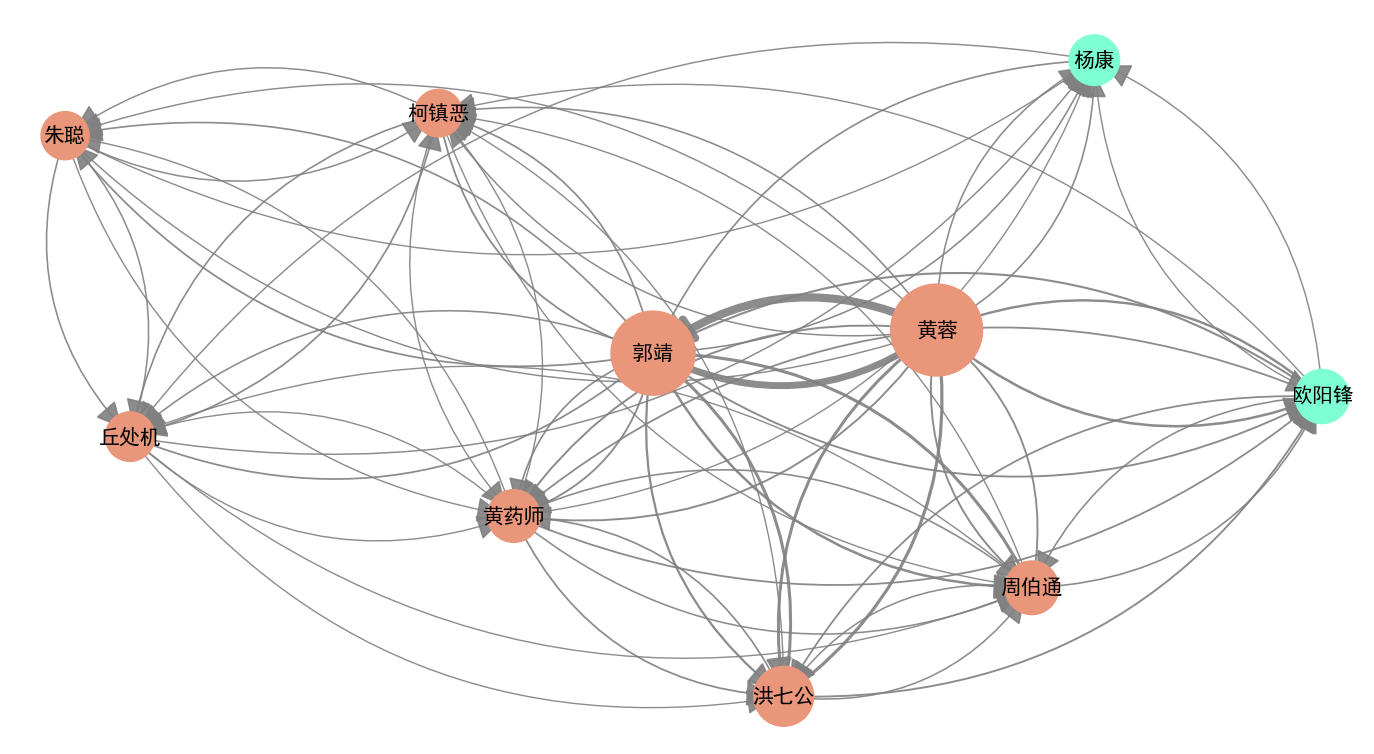}
  \caption{Part of the Character Verbal Relationship Network}
  \label{fig2}
\end{figure}By identifying the participants in novel dialogues, we can easily build character social networks based on dialogue relationships.

We utilize the JY-QuotePlus corpus to construct a social network of character dialogues from the novel \textit{The Legend of the Condor Heroes}. The number of quotations per character is used to determine node size, while the frequency of dialogues between characters is used as the edge weight. The direct edge is from the speaker to the addressee. We select the top 10 characters with the highest number of quotations in the novel to construct the network, and both node size and edge weight are smoothed for better visualization. Furthermore, we set different node colors for the protagonists and villains in the novel to reflect the relationship between characters with different stances. The protagonists are assigned the color darksalmon and the villains are assigned aquamarine. In Figure \hyperref[fig2]{2}, we can observe that the most prominent characters in the network are the male and female protagonists of the novel, “郭靖” (Guo Jing) and “黄蓉” (Huang Rong). Additionally, people closely related to these two people, such as “洪七公” (Hong Qigong) and “黄药师” (Huang Yaoshi), can also be clearly found to be related in the network.

\section{Speaker and Addressee Identification}
\subsection{Methodology}

Different from the traditional approaches, we frame speaker and addressee identification as an extractive reading comprehension task, which identifies the mentioned speakers and addressees directly from the context through understanding the text passages. 
\begin{figure}
  \centering
  \includegraphics[width=\linewidth]{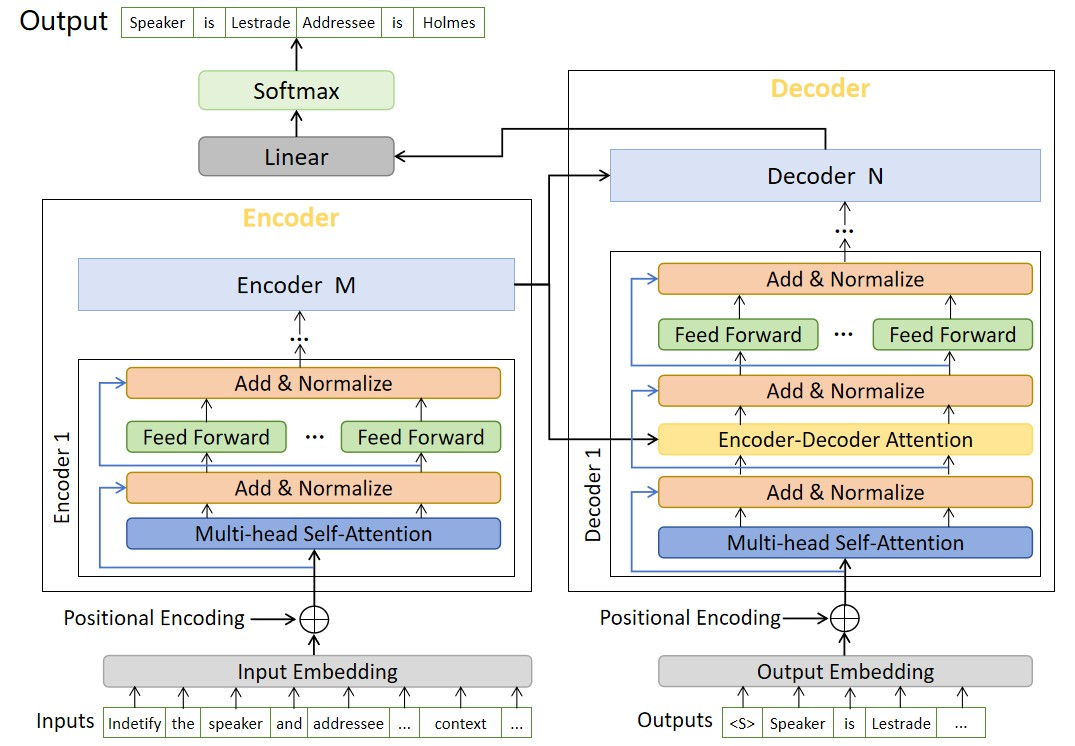}
  \caption{The Prompt Learning based Speaker and Addressee Identification Model}
  \label{fig3}
\end{figure}We adopt a fine-tuning approach based on the T5 (Text-to-Text Transfer Transformer)\cite{raffel2020exploring} model and PromptCLUE for English and Chinese respectively. T5 is a versatile text generation model based on the Transformer architecture, and PromptCLUE is a large-scale, multilingual pre-trained Chinese model supporting learning across diverse NLP tasks.

The overall structure of the proposed model is shown in Figure \hyperref[fig3]{3}. Starting with the pre-trained models, we adapt them to our task by adding a linear layer at the model's output, which is designed for the target language model of speaker and addressee identification. During fine-tuning, we utilize the Adam optimizer to adjust the model parameters. Adam is a widely used gradient descent optimization algorithm that effectively updates model parameters to minimize the loss function, ensuring the model converges to an optimal solution. We also adjust hyperparameters such as learning rate and the number of training epochs to maximize the model's performance.

\subsection{Datasets and Baselines}
Experiments are conducted on two datasets: the self-annotated Chinese JY-QuotePlus and the English RiQua dataset. Details of the JY-QuotePlus dataset are provided in session 3. The RiQua dataset, constructed by Papay et al.\cite{papay2020riqua}, contains 5,963 quotations selected from 11 literary works of 19th-century. Each quotation is annotated with the span of the quotation, the speaker, the addressee, and the speech cues if present. The occurrence rates of speaker and addressee in RiQua are 99.7\% and 95.3\%, respectively.

\begin{figure}
  \centering
  \includegraphics[width=0.9\linewidth]{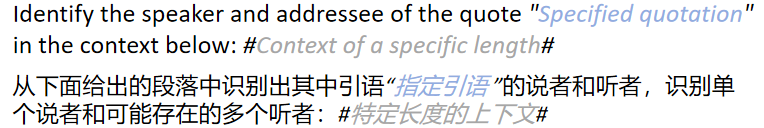}
  \caption{Designed Prompts}
  \label{fig4}
\end{figure}

Prior to the experiments, we perform some preprocessing on the annotated data. Firstly, we process the data labels, selecting quotations that included both speaker and addressee tags and filtering out tags irrelevant to the task, such as cues. Secondly, to ensure the correct execution of the identification task, we design simple prompts tailored to the differing label conditions of the Chinese and English datasets. Figure \hyperref[fig4]{4} shows these prompts. 

Finally, to enable the model to perform the identification task within text passages containing contextual information, we select an appropriate context window. For RiQua, the annotation guidelines and statistical information of the dataset show that the number of speaker and addressee tags in the text preceding a quotation is significantly greater than in the text following it. Therefore, after considering both the text length and the information contained, we ultimately set the context length for the RiQua dataset to 150 tokens before and 30 tokens after each quotation. For JY-QuotePlus, we adopt the format of the original annotated data, selecting five short sentences before and after each quotation to construct the context. 

To validate the effectiveness of our proposed method, we use general large language models with strong text understanding capabilities as baselines, including GPT-3.5\footnote{https://openai.com/} and ChatGLM\footnote{https://chatglm.cn/}. GPT-3.5 is developed by OpenAI, and we use GPT-3.5 Turbo 1106 version. It features enhanced instruction-following capabilities, JSON mode, reproducible outputs, and parallel function calls, among other characteristics. ChatGLM-6B, an open-source conversational language model co-developed by Tsinghua University and Zhipu AI, is based on the General Language Model (GLM) architecture. Its notable advantage is the ability for local deployment on consumer-grade graphics cards.

\subsection{Experiments}
The datasets are divided into training set, validation set and test set in an 8:1:1 ratio. We initially perform multiple rounds of experiments to adjust the model parameters, aiming to select the optimal learning rate and number of training epochs. The final main parameter settings are shown in Table \hyperref[tab3]{3}. We choose accuracy as the evaluation metric. Experiments are run on a NVIDIA 2080 server.

\begin{table}
\centering
\caption{Parameter Settings}
\label{tab3}
\begin{tabular}{>{\centering\arraybackslash}p{4cm} >{\centering\arraybackslash}p{2cm} >{\centering\arraybackslash}p{5cm}}
    \toprule
    Parameters & RiQua - T5 & JY-QuotePlus - PromptCLUE \\
    \toprule
    batch size & 2 & 4 \\
    epochs & 32 & 12 \\
    learning rate & 7e-5 & 8e-5 \\
    maximum text length & 512 & 512 \\
    \bottomrule
\end{tabular}
\end{table}

The experimental results in Table \hyperref[tab4]{4} demonstrate the superior performance of our proposed method based on fine-tuned pre-trained models. The accuracy values of fine-tuned T5 and PromptCLUE in identifying quote speakers and addressees are significantly higher than the large language model baselines. Specifically, the Fine-tuned models achieve an accuracy of 69.60\% in the RiQua dataset and an accuracy of 84.36\% in the JY-QuotePlus dataset for the overall speaker and addressee identification. This reflects the powerful performance of the fine-tuned PTMs in text comprehension and generation.

\begin{table}
\centering
\caption{Accuracy of Speaker and Addressee Identification (\%)}
\label{tab4}
\begin{tabular}{p{1.3cm} >{\centering\arraybackslash}p{1.4cm} >{\centering\arraybackslash}p{1.2cm} >{\centering\arraybackslash}p{1.5cm} >{\centering\arraybackslash}p{1.5cm} >{\centering\arraybackslash}p{1.2cm} >{\centering\arraybackslash}p{1.5cm} >{\centering\arraybackslash}p{1.5cm}}
    \toprule
    \multicolumn{2}{c}{\multirow{2}{*}{Model}} &
    \multicolumn{3}{c}{RiQua} &
    \multicolumn{3}{c}{JY-QuotePlus} \\
    & & speaker & addressee & both & speaker & addressee & both \\
    \midrule
    \multicolumn{2}{c}{Fine-tuned PTMs}
    & 81.13 & 73.58 & \textbf{69.60} & 95.07 & 85.10 & \textbf{84.36} \\
    \midrule
    \multirow{2}{*}{GPT-3.5}
    & zero-shot & 32.34 & 27.24 & 10.90 & 85.47 & 70.32 & 70.07 \\
    & few-shot & 31.81 & 26.36 & 12.83 & 87.56 & 79.93 & 78.82 \\
    \midrule
    \multirow{2}{*}{GLM-6B}
    & zero-shot & 26.42 & 15.51 & 8.81 & 76.60 & 66.01 & 65.64 \\
    & few-shot & 37.74 & 23.69 & 13.21 & 79.56 & 69.83 & 69.09 \\
    \bottomrule
\end{tabular}
\end{table}

\begin{figure}
  \centering
  \includegraphics[width=\linewidth]{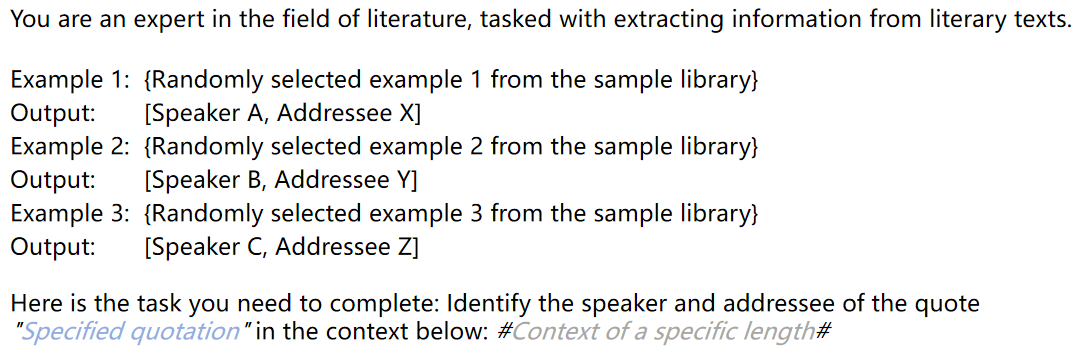}
  \caption{Few-Shot Prompts}
  \label{fig5}
\end{figure}

In contrast, the baseline LLMs in zero-shot setting obtain inferior performance. The primary reason for this discrepancy is that the dataset follows specific annotation guidelines, and without understanding these rules, the model's identified speakers and addressees are significantly different from the original labels. To address this limitation, we also design a few-shot setting for the LLMs. The prompts used in the few-shot setting are shown in Figure \hyperref[fig5]{5}. The experimental results demonstrate that, after being provided with examples, both GPT and GLM exhibit improved performance on the two datasets. After learning from the examples, the LLMs can more accurately identify the speakers and addressees of the quotes.

Additionally, we conduct a novel-wise evaluation on the RiQua dataset to validate fine-tuned PTM's performance on each of the 11 novels that constitute the dataset. In the results presented in Table \hyperref[tab5]{5}, we observe that the identification accuracy varies across different novels. For instance, the novel \textit{The Boscombe Valley Mystery} exhibits the highest identification accuracy, reaching 88.24\%, whereas the accuracy is notably lower for \textit{The Steppe}, standing at 50.75\%. Through our analysis of the dataset, we discover that \textit{The Boscombe Valley Mystery} contains numerous quotations characterized by lengthy textual narrations, where both the speaker and addressee are explicitly mentioned in the surrounding context of the quotation, making the identification task relatively straightforward. Conversely, \textit{The Steppe} features numerous short dialogues between characters, with multiple pronouns referring to different characters within the context paragraphs, thereby introducing certain interference to the model's identification process.
\begin{table}
\centering
\caption{Identification Accuracy of Different Novels (\%)}
\label{tab5}
\begin{tabular}{>{\RaggedRight\arraybackslash}p{6cm} >{\centering\arraybackslash}p{2cm}}
    \toprule
    Author/Novel & Accuracy \\
    \toprule
    Jane Austen	 \\
    \ \ \ 《Emma》 & 68.89 \\
    Charles Dickens	 \\
    \ \ \ 《A Christmas Carol》 & 72.13 \\
    Gustave Flaubert	 \\
    \ \ \ 《Madame Bovary》 & 75.00 \\
    Mark Twain	 \\
    \ \ \ 《The Adventures of Tom Sawyer》 & 64.06 \\
    Sir Arthur Conan Doyle	 \\
    \ \ \ 《A Scandal in Bohemia》 & 68.42 \\
    \ \ \ 《The Red-Headed League》 & 76.92 \\
    \ \ \ 《A Case of Identity》 & 66.67 \\
    \ \ \ 《The Boscombe Valley Mystery》 & \textbf{88.24} \\
    Anton Chekhov  \\
    \ \ \ 《The Steppe》 & 50.75 \\
    \ \ \ 《The Black Monk》 & 76.92 \\
    \ \ \ 《The Lady with the Dog》 & 80.00 \\
    \bottomrule
\end{tabular}
\end{table}

\subsection{Case Study}
We analyze several Chinese and English experimental prediction results. Table \hyperref[tab6]{6} provides three examples. For clarity, we have omitted some excessively long context content in the displayed examples. In the text, the green and blue colors represent the correct speaker and addressee, respectively, while the red color in the labels signifies incorrect predictions.

In the first case, which is particularly challenging, the quotation is spoken by ``he'' (referring to Moisey Moisevitch) to the addressee Kuzmitchov. The fine-tuned T5 model correctly identified both the speaker and the addressee, whereas the baseline model identified the speaker entity Moisey Moisevitch but mistakenly identified the addressee as another present character, Yegorushka. The second case is a classic dialogue pattern from Jin Yong's novel. Only the GLM-6B model made an error, confusing the speaker with the addressee. In the third case, the dialogue structure is unclear, and multiple characters are present. Both the fine-tuned PromptCLUE and GPT models made errors in identification, indicating that the models still struggle with understanding complex text scenarios. Additionally, models sometimes incorrectly identified addressees in multi-character scenes, which is another difficulty that needs further investigation.

\begin{table}
\centering
\caption{Several Prediction Cases}
\label{tab6}
\begin{tabular}{>{\RaggedRight\arraybackslash}p{3.5cm} >{\RaggedRight\arraybackslash}p{8.5cm}}
    \toprule
    \multicolumn{2}{p{12cm}}{Moisey Moisevitch brought a footstool from the other room and sat down a little way from the table. "I wish you a good appetite! Tea and sugar!" he began, trying to entertain his visitors. "I hope you will enjoy it. Such rare guests, such rare ones; it is years since I last saw Father Christopher. And will no one tell me who is this nice little gentleman?" he asked, looking tenderly at Yegorushka. "He is the son of my sister, Olga Ivanovna," answered \textcolor{blue}{\textbf{Kuzmitchov}}. "And where is he going?" "To school. We are taking him to a high school." In his politeness, Moisey Moisevitch put on a look of wonder and wagged his head expressively. \textbf{"Ah, that is a fine thing,"} \textcolor{green}{\textbf{he}} said, shaking his finger at the samovar.} \\
    Ground Truth & Speaker: "he",  Addressee: "Kuzmitchov"  \\
    Fine-tuned T5 & Speaker: "he",  Addressee: "Kuzmitchov"  \\
    GPT-3.5 & Speaker: "Moisey Moisevitch",  Addressee: "\textcolor{red}{Yegorushka}"  \\
    GLM-6B & Speaker: "Moisey Moisevitch",  Addressee: "\textcolor{red}{Yegorushka}" \\
    \toprule
    \multicolumn{2}{p{12cm}}{陆庄主道：“这次来的是那个小王爷的师父，本事可比他大得多，因此我担了心。”\textcolor{green}{\textbf{黄蓉}}道：\textbf{“咦，你怎么知道？”}\textcolor{blue}{\textbf{陆庄主}}道：} \\
    Ground Truth & Speaker: "黄蓉" ,  Addressee： "陆庄主"   \\
    Fine-tuned PromptCLUE & Speaker: "黄蓉" ,  Addressee： "陆庄主"   \\
    GPT-3.5 & Speaker: "黄蓉" ,  Addressee： "陆庄主"   \\
    GLM-6B & Speaker: "\textcolor{red}{陆庄主}" ,  Addressee： "\textcolor{red}{黄蓉}" \\
    \toprule
    \multicolumn{2}{p{12cm}}{\textcolor{blue}{\textbf{裘千仞}}一张老脸一忽儿青，一忽儿白，无地自容，他本想捏造黄药师的死讯，乘乱溜走，哪知自己炫人耳目的手法尽被朱聪拆穿，当即袍袖一拂，转身走出，\textcolor{green}{\textbf{梅超风}}反手抓住，将他往地下摔落，喝道：\textbf{“你说我恩师逝世，到底是真是假？”}这一摔劲力好大，裘千仞痛得哼哼唧唧，半晌说不出话来。黄蓉见那束干茅头上有烧焦了的痕迹，登时省悟，说道：“二师父，你把这束干茅点燃了藏在袖里，然后吸一口，喷一口。”江南六怪对黄蓉本来颇有芥蒂，但此刻齐心对付裘千仞，变成了敌忾同仇。} \\
    Ground Truth & Speaker: "梅超风" ,  Addressee： "裘千仞"   \\
    Fine-tuned PromptCLUE & Speaker: "\textcolor{red}{裘千仞}" ,  Addressee： "\textcolor{red}{梅超风}"   \\
    GPT-3.5 & Speaker: "\textcolor{red}{裘千仞}" ,  Addressee： "\textcolor{red}{黄蓉、江南六怪、朱聪}"   \\
    GLM-6B & Speaker: "梅超风" ,  Addressee： "裘千仞"  \\
    \bottomrule
\end{tabular}
\end{table}


\section{Conclusion} 
This paper aims at automatically identifying speakers and addressees of quotations in novels. In response to the lack of annotated data, we construct the first Chinese novel quotation corpus with elements including speaker, addressee, speaking mode and linguistic cue. We explore prompt learning-based speaker and addressee identification methods with fine-tuned pre-trained models and large language models. Experimental results on both Chinese and English 
corpora validate the effectiveness of the proposed methods. 

In the future, we will study more effective and general approaches for identifying speakers and addressees of quotations to achieve better performance across various datasets. We plan to expand the Chinese quotation corpus to include works from different authors and novels, thereby creating a more diverse dataset. Additionally, we will investigate fictional character relationship analysis based on the extracted speakers and addressees.

\begin{credits}
\subsubsection{\discintname} The authors have no competing interests to declare that are relevant to the content of this article.
\end{credits}

\bibliography{myref.bib} 
	
\bibliographystyle{splncs04} 

\end{CJK*}
\end{document}